\documentclass[10pt,twocolumn,letterpaper]{article}

\usepackage{cvpr}              

\usepackage{graphicx}
\usepackage{amsmath}
\usepackage{amssymb}
\usepackage{booktabs}
\usepackage[accsupp]{axessibility}

\usepackage[pagebackref,breaklinks,colorlinks]{hyperref}

\usepackage[capitalize]{cleveref}
\crefname{section}{Sec.}{Secs.}
\Crefname{section}{Section}{Sections}
\Crefname{table}{Table}{Tables}
\crefname{table}{Tab.}{Tabs.}

\newcommand{\minisection}[1]{\vspace{0.00in} \noindent {\bf #1}\ \ }

\def\cvprPaperID{42} 
\def\confName{CVPR}
\def\confYear{2023}

\begin{document}

\title{Density Map Distillation for Incremental Object Counting}

\author{Chenshen Wu\\
Computer Vision Center\\
Barcelona, Spain\\
{\tt\small chenshen@cvc.uab.es}
\and
Joost van de Weijer\\
Computer Vision Center\\
Barcelona, Spain\\
{\tt\small joost@cvc.uab.es}
}
\maketitle

\begin{abstract}
We investigate the problem of incremental learning for object counting, where a method must learn to count a variety of object classes from a sequence of datasets. A naïve approach to incremental object counting would suffer from \emph{catastrophic forgetting}, where it would suffer from a dramatic performance drop on previous tasks. In this paper, we propose a new exemplar-free functional regularization method, called Density Map Distillation (DMD). During training, we introduce a new counter head for each task and introduce a distillation loss to prevent forgetting of previous tasks. Additionally, we introduce a cross-task adaptor that projects the features of the current backbone to the previous backbone. This projector allows for the learning of new features while the backbone retains the relevant features for previous tasks. Finally, we set up experiments of incremental learning for counting new objects. Results confirm that our method greatly reduces catastrophic forgetting and outperforms existing methods.
\end{abstract}

\section{Introduction} \label{sec:count_introduction}
Image-based counting aims to infer the number of people, vehicles or any other objects present in images. It has a wide range of applications such as traffic control, environment survey and public safety.
Most of existing research focus on learning a model from a single dataset. Only \cite{dkpnet2021chen} and \cite{scalealign2021ma} propose to train a model on multiple datasets simultaneously in a multi-task setting. 
In this paper, we propose a method to incrementally learn to count new objects or to count in a new domain. This has the advantage that it does not require collecting data on a single server for training. Moreover, annotators can focus on just labelling instances of a single class (typically annotated with a single point), which reduces the annotation effort required for adding new classes.

Continual learning (CL) addresses the problem of training a model from a non-stationary distribution. It is important because data in the real-world might not be jointly available (e.g. due to data privacy or legislation). Moreover, often the previous data cannot be revisited due to the privacy or storage restrictions. Researchers have explored continual learning in many tasks, e.g. classification~\cite{lwf2017li,ewc2017kirkpatrick}, segmentation~\cite{mib2020Cermelli, MAML2017Finn}, and object detection~\cite{Shmelkov2017ILOD}. However, continual learning for counting systems, has to the best of our knowledge, not 
 yet been studied.

\begin{figure*}[ht]
\setlength{\tabcolsep}{0pt}

\begin{center}
\begin{tabular}{cccc}
\includegraphics[width=0.23\textwidth]{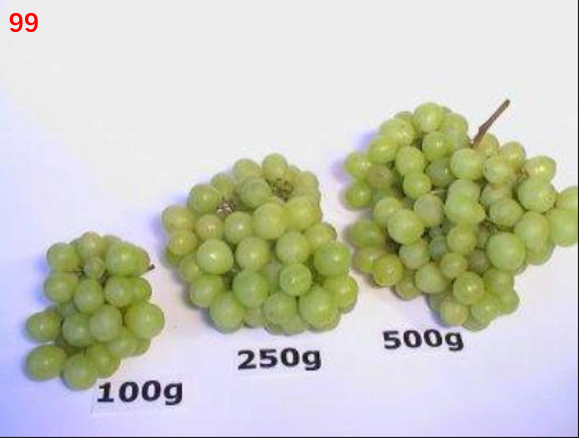} &
\includegraphics[width=0.23\textwidth]{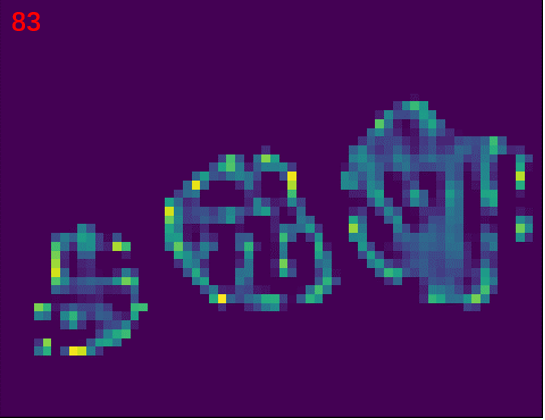} &
\includegraphics[width=0.23\textwidth]{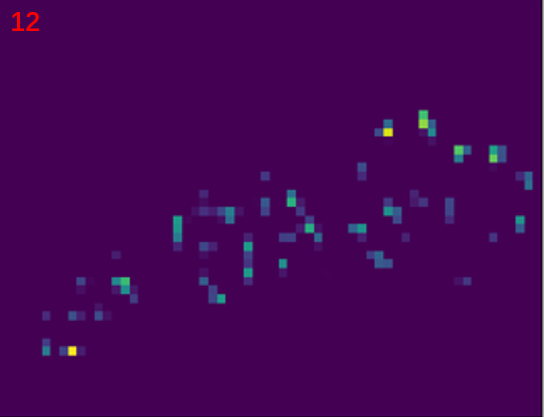} &
\includegraphics[width=0.23\textwidth]{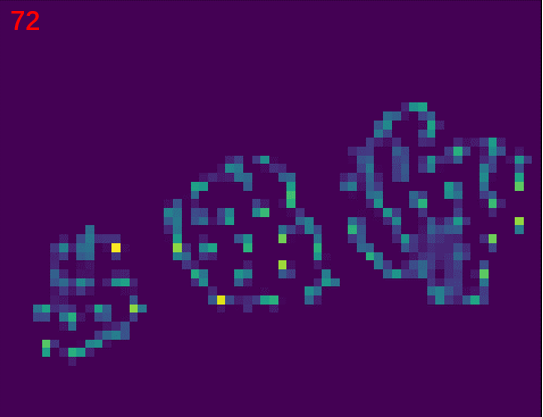} \\
(a)   & (b) & (c)  & (d) 
\end{tabular}
\end{center}
\caption{(a) Input Image (b) Density map after training only on the \emph{grapes} data. Density map after learning two  additional tasks (\emph{tomatoes} and \emph{strawberries}) with (c) Fine Tuning (FT) and (d) after learning by our proposed method (DMD). In the upper-left corner we show the ground truth number of grapes in (a) and the estimation of the algorithms respectively. Note that naive fine-tuning leads to catastrophic forgetting and the method loses its ability to count \emph{grapes}. (d) Our method manages to get a considerable better count prediction even though there is some performance loss.}
\label{fig:counting_catastrophic_forgetting}
\end{figure*}
One of the main challenges of continual learning is catastrophic forgetting. After training on new data, models tend to forget the knowledge extracted from previous data. In the past few years, people tried to alleviate this issue by using replay examples~\cite{rebuffi2017icarl, bic2019}, expand networks~\cite{yoon2017DynamicExpand} and regularization~\cite{ewc2017kirkpatrick, lwf2017li}. As one of the most promising methods, regularization can be further categorized as weight regularization~\cite{ewc2017kirkpatrick} and data (or functional) regularization~\cite{lwf2017li}. The former applies regularization on weights to prevent them from drifting too far from the old model, while the latter apply it on the output of the network  given the input data. Due to their success for the classification tasks~\cite{masana2020CIL}, the fact that they do not require exemplars, and because they scale well with the number of tasks, we will here explore data regularization for object counting.

However, these methods are mainly designed for the classification problem, which aims to predict a category for a given sample. For the counting problem, which is a regression problem where the output is a scalar map, we found that directly applying these existing CL method is suboptimal. Therefore, we propose a new method called \emph{Density Map Distillation} (DMD).
For each new object, we train a separate counter head that maps the feature extraction backbone to an object-specific density map.
After the training of each task, the counter head is fixed and during future tasks only the feature extractor and new counter head is trained. 
When training a new task, we use the new data to apply distillation on all previous counter heads.
Since the feature extractor is drifting when learning new tasks, we propose to use a cross-task adaptor to project the new features to the old features. This mechanism allows us to keep plasticity while maintaining stability (i.e., prevent forgetting).

The contributions of this paper include: (1) We set up experiments for incremental learning for counting new objects. We define metrics for evaluating incremental counting problems. 
(2) We propose Density Map Distillation (DMD) for the incremental counting problem. The method includes fixing the task-specific counter head and training a cross-task adaptor for the feature extractor. Our method prevents forgetting, while maintaining plasticity to learn new tasks. 
(3) We adapt several existing methods of incremental learning for incremental object counting. Experiments show that our new methods outperforms these existing methods.

\section{Related Work}

\subsection{Incremental Learning}
Incremental learning aims to develop methods that can learn new knowledge from new data while not forgetting previous knowledge learned from the previous training stages. 
The existing methods can mainly be categorized as three types: distillation based, dynamic model based and rehearsal based~\cite{delange2021survey_TIL}. 
Distillation based methods focus on how to limit the change of the model by applying a loss on the weights directly~\cite{ewc2017kirkpatrick, mas2018},  or on the output features~\cite{lucir2019} and probabilities~\cite{lwf2017li}.  Dynamic model based methods~\cite{yoon2017DynamicExpand} extend the architecture of the network to learn new knowledge from the new incoming data distribution. Rehearsal based methods~\cite{rebuffi2017icarl} save a few exemplars from the previous dataset and replay them or use them to constrain the model during the new training sessions.  

Previously, incremental learning mainly focused on image classification problems. Recently, the community also developed incremental learning algorithms for other problems such as image generation\cite{mergan2018wu}, segmentation\cite{mib2020Cermelli}, object detection\cite{objectfewshotIL2020}, video classification\cite{Park2021VideoIL}. But to the best of our knowledge, there is no work for incremental learning of counting problems yet.

\subsection{Crowd Counting}
There are two categories of crowd counting methods, density map based methods\cite{bayesianloss2019Ma, dmcount2020wang} and localization based methods\cite{p2pnet2021Song}. 

Localization based methods count by locating each individual's position. Some methods \cite{LSCCNN20, Lian2019DetectionRGBD} are driven by an object detector, and inaccuracy is introduced by estimation of the ground truth bounding boxes. 
Liu et al.~\cite{Liu2019Zooming}  propose RAZ-Net that recurrently detect high density regions and zoom in for re-inspection. The network performs the counting and localization task at the same time, and these two tasks complement each other. 
Song et al.~\cite{p2pnet2021Song} propose the P2PNet that predicts the localization points directly by introducing a one-to-one matching strategy from the prediction to the ground truth. 

For the localization based method, it is hard to predict each location where the crowd density is very high~\cite{Wan2021GeneralizedLoss}. Most of the research in counting mainly focuses on predicting a density map and then count by the summing it. 
In~\cite{MultiColum2016Zhang, Sam2017SwitchingCNN}, they propose to use several parallel CNNs of different sizes to address the problem of scale variation. Another line of research focuses on the loss function. In~\cite{bayesianloss2019Ma}, Ma et al. propose a Bayesian loss to measure the distance between the predicted and the ground truth density map. 
Wang et al.~\cite{dmcount2020wang} measure the similarity between the predicted density map to the ground truth density map by solving an Optimal Transport (OT) problem.

In the above methods, the model is always trained with one dataset. In \cite{dkpnet2021chen} and \cite{scalealign2021ma}, they propose to train a model on multiple datasets simultaneously. 
Some other researches~\cite{Wang2019CountWild, Han2020CountDA1, Gao2020CountDA2} focus on counting problem in domain adaptation setting, where the model is trained on the source dataset and the label of the target dataset is limited. 
In~\cite{Gao_2023}, Gao et al. deal with the problem of domain incremental learning of crowd counting, where a model is trained to count people sequencially on several datasets. 
In~\cite{Lu2018ClassAgnosticCounting,counteverything2021Ranjan,bmnet2022Shi}, they consider the problem of class-agnostic counting. The aim is to train a network that counts the number of instances in an image by specifying an exemplar patch.

\begin{figure*}[t]
    \centering
    \begin{tabular}{c c}
    \includegraphics[width=0.35\linewidth]{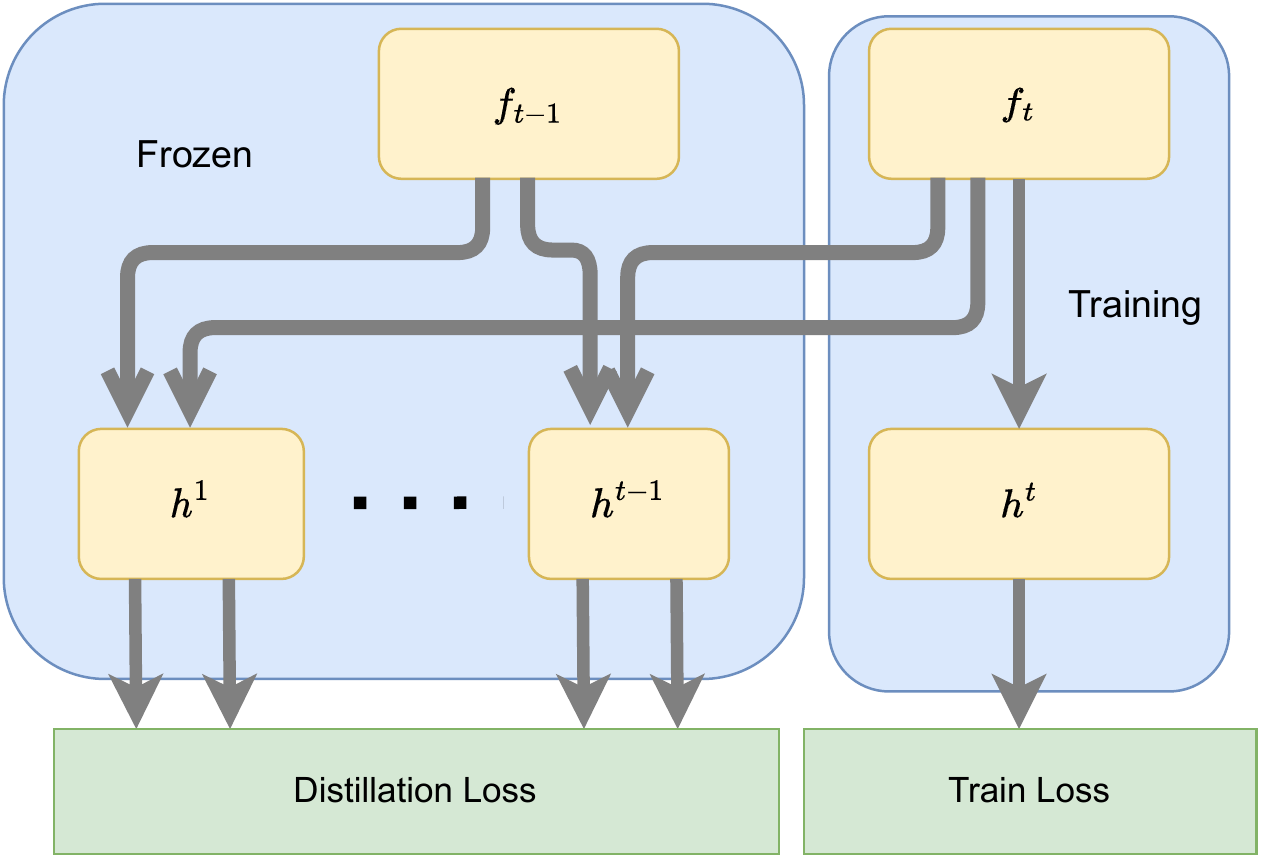} &
    \includegraphics[width=0.48\linewidth]{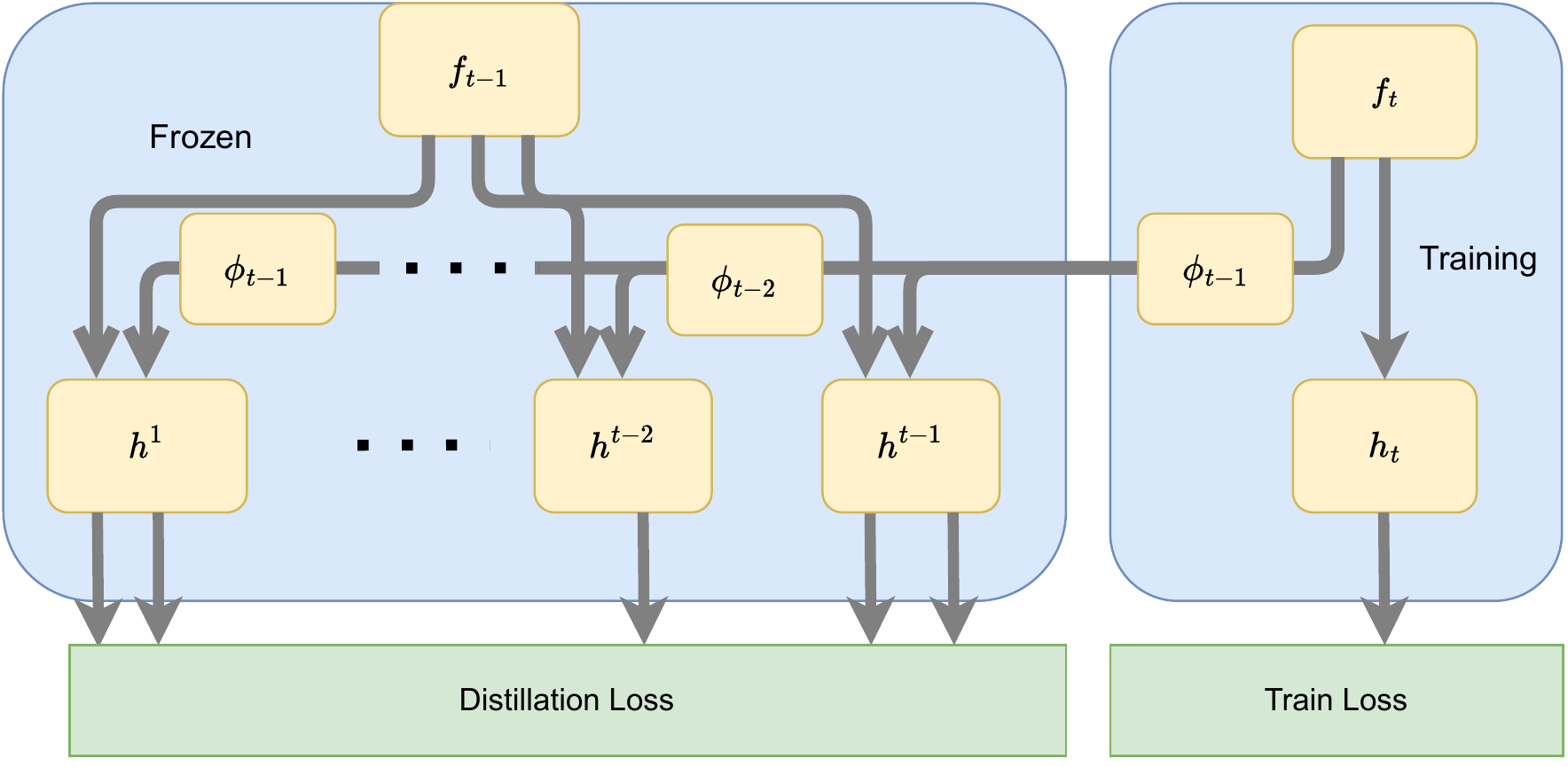} \\
    (a) Density Map Distillation without Adapter & (b) Density Map Distillation
    \end{tabular}
  \caption{ (a) Density Map Distillation (DMD) without Adaptor. While training new tasks, the distillation loss is applied on the output density map using the previous counter heads, between the previous and the new feature extractors. Different from LwF\cite{lwf2017li}, previous counter heads are fixed when training new task. 
    (b) Density Map Distillation (DMD). In addition to the distillation loss, we train cross-task adaptors ($\phi$) to project  new features to old features, since 
    the feature extractor is continuously trained.}
    \label{fig:DMD}
\end{figure*}

\section{Method}
Counting is an integral part of many real-life applications. To alleviate the human costs of manual counting, many methods have been developed for the counting of objects~\cite{bayesianloss2019Ma, dmcount2020wang, p2pnet2021Song}. As discussed in the introduction, these methods generally assume that all training data is jointly available. However, for many applications this assumption is not realistic and the algorithm would only be able to have access to a batch of data at each time step.

A naive approach to learning from a sequence of tasks would be to just continue finetuning the model on the available data of consecutive tasks. However, this would lead to the \emph{catastrophic forgetting} phenomenon. An illustration of this is provided in Figure~\ref{fig:counting_catastrophic_forgetting} where we show that after learning several tasks with fine-tuning, the method has lost its ability to count the first-task \emph{grapes} class. In this section, we explore distillation-based methods for incremental learning of object counting to prevent the effect of catastrophic forgetting.

\subsection{Notation}
In a typical counting problem, images $X_i$ are annotated for a single object class $c \in C$, for example annotations of persons, cars, or apples are given. Existing works do not consider counting various classes of objects simultaneously. Typically, objects are annotated with a single point in the center of the object at positions $p_{ij}, j \in {1,...,N_i}$ where there are $N_i$ objects for the image $x_i$. We will use the notation $P_i$ to refer to the set of locations in image $x_i$. Object counting learns a model for a single object class that given an input image maps to a density map which predicts the number of object instances per pixel \cite{bayesianloss2019Ma, dmcount2020wang} or which directly predicts the object coordinates \cite{p2pnet2021Song}. 

In incremental learning for counting problems, the data is split in various tasks, where each task $t \in [1,T]$ arrives sequentially. For each task, the dataset $D_t=\{c_t, \bigl(\left(x_1,P_1\right),\left(x_2,P_2\right),\cdots,\left(x_M,P_M\right)\bigr)\}$ contains the class category $c_t$ and images with ground truth position annotations. We consider the scenarios where each task has a single object category $c_t$ different from the other tasks. After training on all $T$ tasks, the model is evaluated on a test set $Y$ that contains images of all objects $C$ seen in the various tasks. The task-ID of the test images is available to the algorithm at inference time (this setting is also known as task-incremental learning)~\cite{ThreeScenarios2019VandeVen}. 

For the training of the object counting network, we propose to use a network which can be divided into a feature extractor $f : R^{w\times h \times 3}\rightarrow R^{w_d \times 
 h_d \times d}$ where $d$ is the number of output channels of the feature extractor, and an object-specific \emph{counter head} given by $h: R^{w_d \times 
 h_d \times d} \rightarrow R^{w_d \times 
 h_d \times 1}$. The counter head maps from the feature space to a density map. The prediction of a network for an image $x$ is then given by:
\begin{equation}
    \hat y=\sum_{w=1}^{w_d}\sum_{h=1}^{h_d} \hat d(x) = \sum_{w=1}^{w_d}\sum_{h=1}^{h_d} h\circ f(x)
\end{equation}
where $\hat d=h\circ f$ is the predicted density map and the summation is over the spatial coordinates of the density map.

For training the new task, we use the loss proposed by Wang et al. \cite{dmcount2020wang}:
\begin{align}
    \mathcal{L}_\text{train} = \left| \left\Vert d \right\Vert_1 - \left\Vert \hat d \right\Vert_1 \right| - \lambda_1 \mathcal{W} \left( \frac{d}{\left\Vert d \right\Vert_1} - \frac{\hat d}{\left\Vert \hat d \right\Vert_1} \right) \notag\\ +  \lambda_2 \frac{1}{2} \left \Vert \frac{d}{\left \Vert d \right\Vert_1} - \frac{\hat d}{\left \Vert \hat d \right \Vert_1} \right \Vert 
\end{align}
The first term is the counting loss for the final counting number. The second term is the optimal transport loss, where $\mathcal{W}$ is the Monge-Kantorovich's Optimal Transport (OT) cost\cite{villani2008optimal}. The third term is the Total Variation (TV) loss, and $\lambda_1$ and $\lambda_2$ are the hyperparameters for the OT and TV losses. 

To extend the above described method to incremental object counting, we use the following notations. The network contains a feature extractor after learning task $t$ given by $f_t$. For each  of the learned tasks, we have a task specific counter head $h_t$ for each object.
At the beginning of the task, the feature extractor $f_t$ is initialized from the previous feature extractor $f_{t-1}$. The previous feature extractor $f_{t-1}$ is then fixed and stored. Other older feature extractors like $f_{t-2}$ are not stored. 

When training task $t$ we use $h_t^\tau$ to refer to the previous counter heads for the object that was learned at task $\tau$. At inference time, we combine the last feature extractor with any of the previously learned counter heads, so for example to get the solution for class $c_\tau$ after training task $t$ we apply $h^{\tau}_t\circ f_t$. We also consider fixing the previous task specific counter, i.e. we do not update it when learning new tasks, so $h_\tau^\tau=h_{\tau+1}^\tau=\cdots$, and we simply refer to it as $h^\tau$.

\subsection{Data regularization for regression problems}
One of the popular approaches to prevent \emph{catastrophic forgetting} in continual learning is by means of regularization methods~\cite{delange2021survey_TIL}. Compared with the other two main approaches to continual learning, regularization methods have the advantage over rehearsal methods that they do not require the storage of any data from previous tasks, and they do not have an increased memory footprint when training on larger task sequences like isolation methods typically have. Regularization methods can be differentiated in data (or functional) and weight regularization methods. 

Data regularization for classification networks is proposed by \cite{lwf2017li} and it is one of the most popular methods for exemplar-free continual learning. Different from the weight regularization methods \cite{ewc2017kirkpatrick, mas2018} which apply the regularization loss on the parameters of the network, data regularization apply the regularization on the output of network layers. Other than weight regularization, it is dependent upon the data on which the distillation is applied. This idea has been further extended by \cite{lucir2019, podnet2020}. The former apply the regularization loss on the feature output and the output after the cosine normalization. The latter apply them on several intermediate layers and study various marginalizations to improve the plasticity of the method.

\begin{figure*}[t]
    \centering
    \begin{tabular}{c c c c}
    \includegraphics[width=0.2\textwidth]{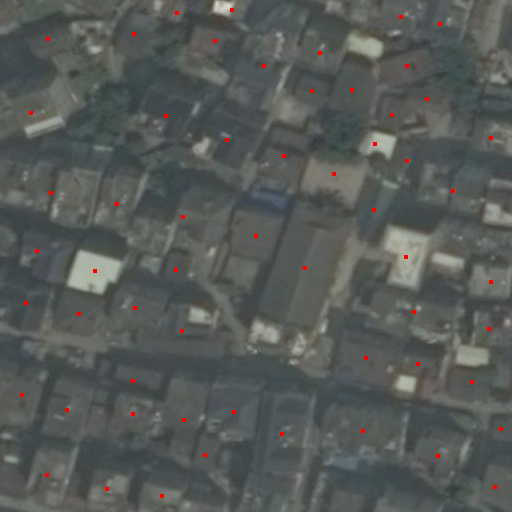} &
    \includegraphics[width=0.2\textwidth]{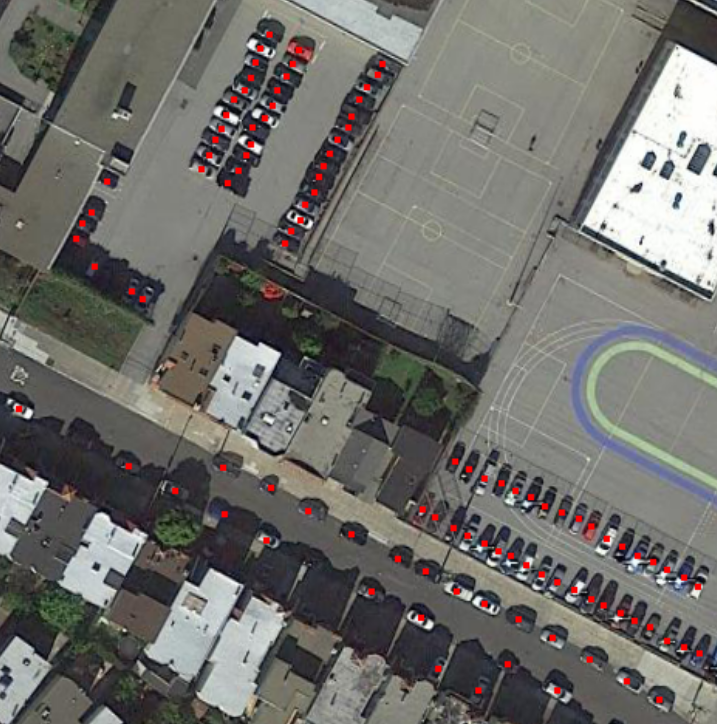} &
    \includegraphics[width=0.23\textwidth]{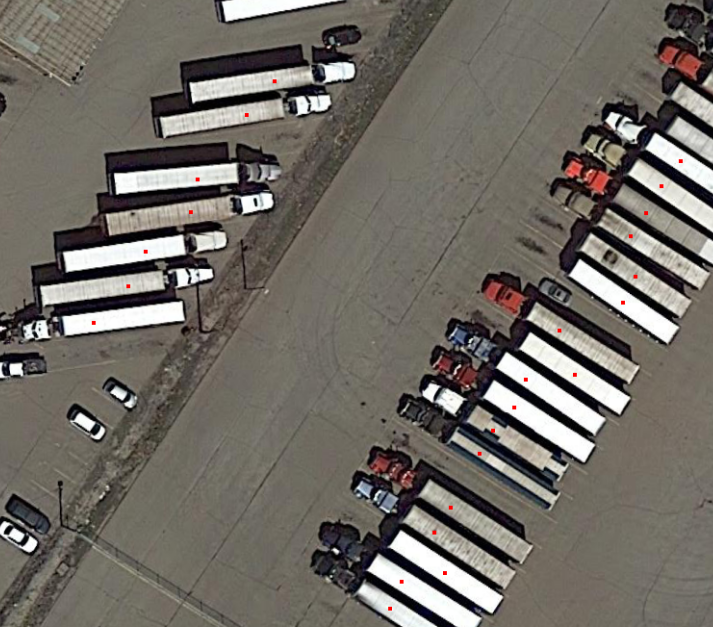} &
    \includegraphics[width=0.24\textwidth]{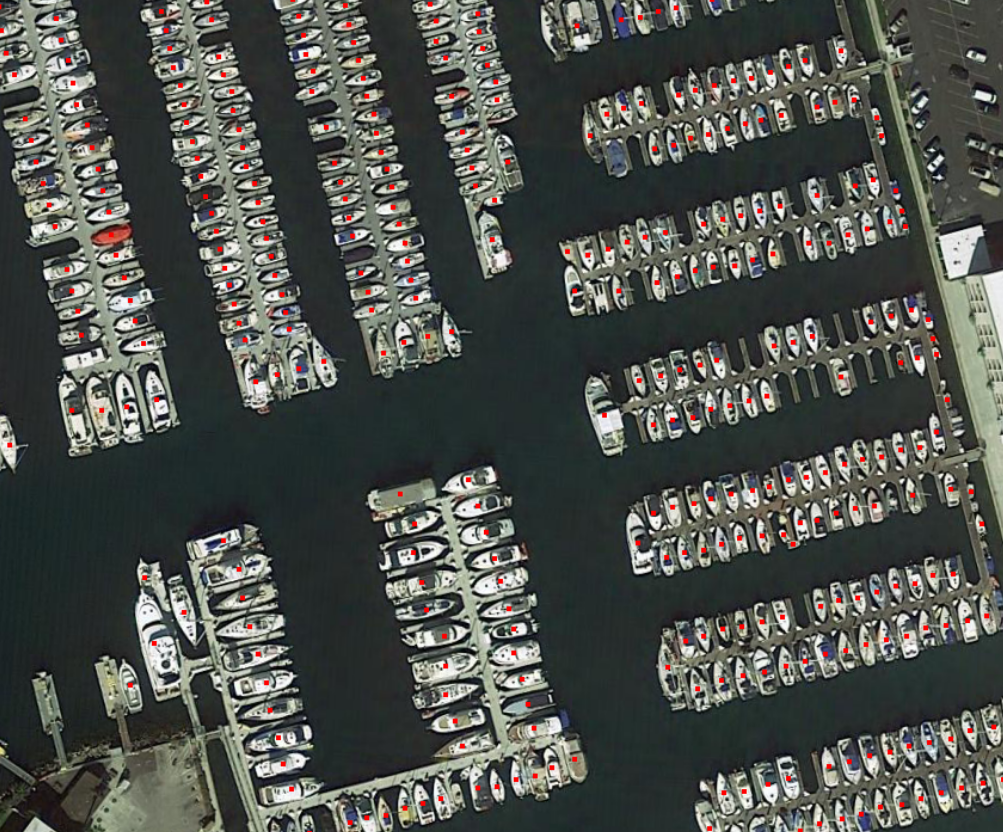} \\
    building & small vehicle & large vehicle & ship
  \end{tabular}
  \caption{{Sample images from RSOC dataset \cite{countingfromsky2020gao}}. }
\end{figure*}

\begin{figure*}[ht]
    \centering
    \begin{tabular}{c c c c}
    \includegraphics[width=0.21\textwidth]{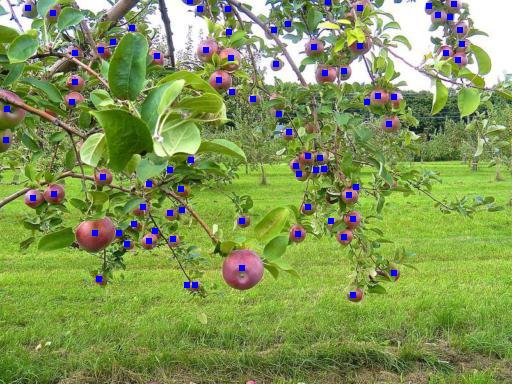} &
    \includegraphics[width=0.17\textwidth]{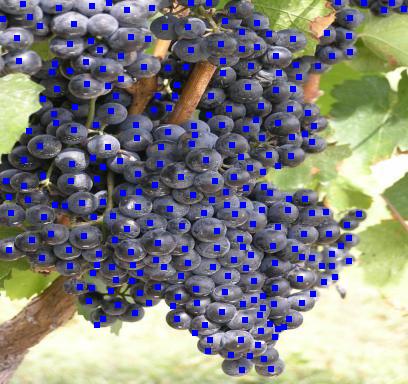} &
    \includegraphics[width=0.235\textwidth]{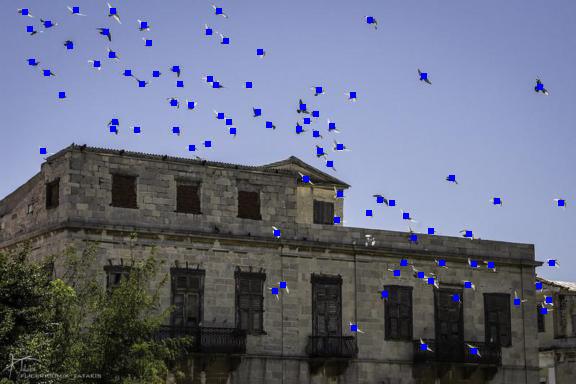} &
    \includegraphics[width=0.23\textwidth]{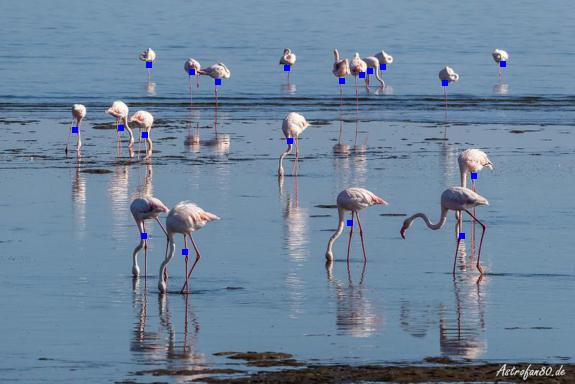} \\
    apples & grapes & pigeons & flamingos
  \end{tabular}
  \caption{{Sample images from FSC147 dataset \cite{counteverything2021Ranjan}}. }
\end{figure*}

However, the most popular data regularization method, LwF\cite{lwf2017li}, cannot be applied directly to the counting problem. In LwF\cite{lwf2017li}, Li et al. proposed to apply a knowledge distillation loss between the new and the old output. Given the image from the new dataset as the input, both models give a prediction of the probability and a cross entropy loss is applied as a regularization. However, an object counting network does not output a probability, and therefore the cross entropy loss cannot be applied. An adaption to the counting problem is to apply a $L_2$ loss on the density map:
\begin{equation}
    \mathcal{L}_\text{reg} = \sum_{\tau \in [1,t-1]}  \left\Vert h_t^\tau \circ f_t(x)- h_{t-1}^\tau \circ f_{t-1}(x)\right\Vert_2.
    \label{eq:counting_lwf}
\end{equation} 
We will identify this method with Learning without Forgetting (LwF) in our results section. However, we found that such an adaptation leads to suboptimal results. We hypothesize that this method suffers from overfitting.

Another typical data regularization method is to apply regularization on the feature level according to: 
\begin{equation}
    \mathcal{L}_\text{reg}=\sum_{\tau \in [1,t-1]}  \left\Vert f_t(x) - f_{t-1}(x)\right\Vert_2.
\end{equation} 
We call this method Feature Distillation (FD)~\cite{lucir2019, podnet2020}. This prevents the feature extractor from drifting too far from the old one. This regularization is very restrictive since it requires the whole feature  map to remain similar. So it is often too rigid so that the model cannot learn from new tasks. This was also observed by PODNet~\cite{podnet2020}. 

\subsection{Density Map Regularization with Cross-Task adaptors}
To address the shortcomings of data regularization for regression tasks, and to prevent the overfitting of the previous counter heads, we propose a further adaptation. 
After training of each task, the counter head for this task will be fixed. So the notation $h_t^\tau$ (the counter head for task $\tau$ during or after the learning of the task $t$) can be simplified as $h^\tau$, because the counter head is not changed after the training, and hence $h_\tau^\tau=h_{\tau+1}^\tau=\cdots$. We also store the previous feature extractor $f_{t-1}$ as a reference for the regularization loss. Earlier feature extractor are not needed, so the memory requirement does not scale linearly with the number of tasks. Then we apply the following regularization loss on the density map output from the old and new models: 
\begin{equation}
    \mathcal{L}_\text{reg}=\sum_{\tau \in [1,t-1]}  \left\Vert h^\tau \circ f_t(x) - h^\tau \circ f_{t-1}(x)\right\Vert_2.
    \label{eq_DMD-adapt}
\end{equation}
This method is an exemplar-free method, since images from previous tasks are not used. As shown in Figure~\ref{fig:DMD}.a, both old and new feature extractors use the same image $x \in D_t$ as input and extract a feature $f_{t-1}(x)$ and $f_t(x)$. We use $L_2$ distance as the regularization loss. It is applied to the output of each counter head for all previous tasks $h_1,\cdots,h_{t-1}$, which encourages the new model output to yield the same result when counting previous objects.

\begin{table*}[tb]
\begin{center}
\scalebox{0.6}{
\begin{tabular}{|c|ccc|ccc|ccc|ccc|c|}
\hline

\textbf{Dataset:} & \multicolumn{3}{|c|}{\textbf{building}}  & \multicolumn{3}{|c|}{\textbf{small vehicle}} & \multicolumn{3}{|c|}{\textbf{large vehicle}}& \multicolumn{3}{|c|}{\textbf{ship}} & \textbf{Avg} \\

\textbf{Metric:} & MSE & MAE & NMAE & MSE & MAE & NMAE & MSE & MAE & NMAE & MSE & MAE & NMAE & NMAE \\
\hline
FT & 31.10 & 27.05 & 0.926 & 1266.89 & 430.81 & 0.609 & 52.34 & 38.70 & 0.624 & 86.40 & 59.51 & 0.296 & 0.614 \\
\hline
LwF & 14.21 & 10.70 & 0.360 & 1326.12 & 505.10 & 1.000 & 79.65 & 62.78 & 1.000 & 137.76 & 108.27 & 0.499 & 0.715 \\
FD & \bf{10.79} & \bf{7.48} & \bf{0.263} & 1108.06 & 356.01 & 0.379 & 39.16 & 27.17 & 0.423 & 122.86 & 88.16 & 0.382 & 0.362 \\
EWC & 10.94 & 7.58 & 0.268 & 1150.12 & 360.78 & 0.383 & 39.75 & 27.33 & 0.418 & 117.46 & 80.90 & 0.345 & 0.352 \\
MAS	& 11.07 & 7.71 & 0.271 & 1068.67 & 333.95 & 0.380 & 40.23 & 27.75 & 0.419 & 117.94 & 85.17 & 0.375 & 0.361 \\
DMD w/o Adapt & 13.36 & 9.90 & 0.336 & \bf{929.75} & \bf{291.62} & \bf{0.288} & 33.92 & 22.52 & 0.345 & 130.56 & 87.13 & 0.384 & 0.338 \\
DMD & 12.63 & 9.25 & 0.315 & 988.63 & 320.97 & 0.315 & \bf{25.78} & \bf{16.53} & \bf{0.269} & \bf{107.84} & \bf{76.40} & \bf{0.367} & \bf{0.316} \\

\hline

\end{tabular}
}
\end{center}
\caption{Performance of several incremental learning methods after learning four tasks of RSOC dataset. In {\bf bold} we show the best results for each column excluding the FT method.}
\label{tab:counting_table_rsoc}
\end{table*}

\begin{table*}[tb]
\begin{center}
\scalebox{0.6}{
\begin{tabular}{|c|ccc|ccc|ccc|ccc|c|}
\hline
\multicolumn{14}{|c|}{\textbf{FSC-fruits}} \\
\hline

\textbf{Dataset:} & \multicolumn{3}{|c|}{\textbf{grapes}}  & \multicolumn{3}{|c|}{\textbf{tomatoes}} & \multicolumn{3}{|c|}{\textbf{strawberries}}& \multicolumn{3}{|c|}{\textbf{apples}} & \textbf{Avg} \\

\textbf{Metric:} & MSE & MAE & NMAE & MSE & MAE & NMAE & MSE & MAE & NMAE & MSE & MAE & NMAE & NMAE \\
\hline
FT & 67.11 & 53.25 & 0.671 & 28.49 & 20.28 & 0.362 & 28.11 & 21.18 & 0.312 & 12.85 & 7.89 & 0.119 & 0.366 \\
\hline
LwF & 27.29 & 19.74 & 0.260 & 63.09 & 49.54 & 1.000 & 73.45 & 61.02 & 1.000 & \bf{16.50} & \bf{9.77} & \bf{0.135} & 0.599 \\
FD & 17.26 & 11.99 & 0.149 & 16.44 & 12.76 & 0.319 & 19.83 & 13.79 & 0.225 & 29.35 & 15.19 & 0.179 & 0.218 \\
EWC & 17.91 & 12.40 & 0.154 & 17.86 & 14.07 & 0.328 & 21.42 & 15.16 & 0.252 & 22.03 & 13.69 & 0.190 & 0.231 \\
MAS & \bf{16.12} & \bf{11.37} & \bf{0.142} & 15.79 & 12.19 & 0.298 & 19.23 & 13.31 & 0.212 & 27.32 & 15.61 & 0.211 & 0.216 \\
DMD w/o Adapt & 25.35 & 18.28 & 0.240 & 14.29 & 11.50 & 0.260 & \bf{16.09} & 11.25 & \bf{0.177} & 23.96 & 12.33 & 0.141 & 0.207 \\
DMD & 26.63 & 18.83 & 0.250 & \bf{11.35} & \bf{8.86} & \bf{0.221} & 16.27 & \bf{11.09} & 0.178 & 18.56 & 10.84 & 0.140 & \bf{0.197} \\

\hline
\multicolumn{14}{|c|}{\textbf{FSC-birds}} \\
\hline

\textbf{Dataset:} & \multicolumn{3}{|c|}{\textbf{flamingos}}  & \multicolumn{3}{|c|}{\textbf{pigeons}} & \multicolumn{3}{|c|}{\textbf{cranes}} & \multicolumn{3}{|c|}{\textbf{geese}} & \textbf{Avg} \\

\textbf{Metric:} & MSE & MAE & NMAE & MSE & MAE & NMAE & MSE & MAE & NMAE & MSE & MAE & NMAE & NMAE \\
\hline
FT & 74.55 & 37.77 & 0.587 & 41.54 & 27.12 & 0.617 & 12.79 & 7.63 & 0.218 & 7.90 & 3.54 & 0.102 & 0.381 \\
\hline
LWF & 66.35 & 27.73 & 0.326 & 60.12 & 42.93 & 1.000 & 54.75 & 32.30 & 1.000 & 10.59 & 5.21 & 0.153 & 0.620 \\
FD & 64.01 & 24.85 & 0.246 & 27.42 & 15.30 & 0.350 & 13.14 & 7.09 & 0.192 & 13.19 & 6.78 & 0.198 & 0.247 \\
EWC & \bf{62.90} & \bf{23.94} & 0.233 & 23.33 & 12.06 & 0.284 & 12.76 & 6.49 & 0.163 & 12.25 & 6.41 & 0.193 & 0.217 \\
MAS & 63.38 & 24.10 & \bf{0.226} & 25.07 & 13.70 & 0.308 & 12.96 & 6.64 & 0.168 & 12.04 & 6.03 & 0.178 & 0.220 \\
DMD w/o Adapt & 67.35 & 28.19 & 0.330 & 25.96 & 11.60 & 0.204 & 7.78 & 4.41 & 0.128 & 10.72 & 5.56 & 0.166 & 0.207 \\
DMD & 67.21 & 28.66 & 0.354 & \bf{22.52} & \bf{10.56} & \bf{0.198} & \bf{7.23} & \bf{4.16} & \bf{0.123} & \bf{9.01} & \bf{4.37} & \bf{0.133} & \bf{0.202} \\
\hline

\end{tabular}
}
\end{center}
\caption{Performance of several incremental learning methods after learning four tasks on FSC-fruits and FSC-birds. In {\bf bold} we show the best results for each column excluding the FT method.}
\label{tab:counting_table_FSC}
\end{table*}

As we fixed the previous counter head, this might prevent the feature extractor from learning new knowledge. Therefore, in addition, we propose to train an adaptor $\phi$ to project the features from the new feature extractor to the old one. The adaptor is trained together with the feature extractor using the distillation loss. 
As illustrated in Figure~\ref{fig:DMD}.b, when training task $t$, the adaptor $\phi_{t-1}$ projects the features generated by $f_t$ to approximate those by $f_{t-1}$. Similarly, by cascading several previous adaptors $ \phi_{t-2},\cdots,\phi_1 $, the features can be projected to those in earlier stages. So the distillation loss with adaptor is given by:
\begin{align}
    \mathcal{L}_\text{reg}=\sum_{\tau \in [1,t-1]}  \Vert h^\tau \circ \phi_\tau \circ \cdots \circ \phi_{t-1} \circ  f_t(x) \notag\\ -  h^\tau \circ \phi_\tau \circ \cdots \circ \phi_{t-2} \circ  f_{t-1}(x) \Vert_2.
    \label{eq:DMD}
\end{align}
We call our method \emph{density map distillation} (DMD). To identify, the version defined by Eq.~\ref{eq_DMD-adapt} without the adaptor, we will use the name \emph{DMD w/o Adapt}). 

The final loss is given by: 
\begin{equation}
    \mathcal{L} = \mathcal{L}_\text{train} + \lambda \mathcal{L}_\text{reg},
\end{equation}
where $\lambda$ is the hyperparameter to balance the training loss and the regularization loss.

During the inference, the feature is extract by the new feature extractor $f_t$. To count the object $c_\tau$, the feature needs to be adapted through all the adaptors learned after that task, $\phi_{t-1},\phi_{t-2},\cdots,\phi_\tau$. Then counter head $h^\tau$ uses the adapted feature to predict the density map for the given object according to: 
\begin{equation}
    \hat d(x) = h^\tau \circ \phi_\tau \circ \cdots \circ \phi_{t-1} \circ  f_t(x).
\end{equation}

The learning of adaptors between backbone networks in continual learning has been studied recently for continual self-supervised learning~\cite{gomez2022continually, fini2022self}, and more recently for supervised tasks in~\cite{cotogni2022gated}. Other than them, we here 
study their usage for a regression task. 

\section{Experimental Results}

\subsection{Dataset and evaluation}
\paragraph{Datasets.} The RSOC dataset is a counting dataset of aerial images proposed by \cite{countingfromsky2020gao} involving \emph{buildings}, \emph{small vehicles}, \emph{large vehicles}, and \emph{ships}. In this paper, we will consider learning to count these classes incrementally in the before mentioned order. The images of buildings are collected from Google Earth, while the rest are from the DOTA dataset~\cite{dota2018xia}. The DOTA dataset is an object detection dataset of aerial images. The original labels of bounding boxes are replaced by their central location for the counting problem. There are 2468 images for buildings, 280 images for small vehicles, 172 images for large vehicles and 137 images for ships

The FSC147\cite{counteverything2021Ranjan} dataset is a counting dataset for few-shot learning, containing 147 categories. For most categories, there are less than 100 images per category. For our incremental learning, we chose several categories that contain a significant number of images. To better share the knowledge across the learning process, we select similar categories for the learning sequence. We consider two sequences of counting of four tasks. 
The first sequence, called \emph{FSC-fruits}, contains \emph{grapes}, \emph{tomatoes}, \emph{strawberries} and \emph{apples}, containing 116, 117, 126 and 165 images, respectively.
The second sequence, called \emph{FSC-birds}, is \emph{flamingos}, \emph{pigeons}, \emph{cranes} and \emph{geese}, containing 76, 81, 108 and 162 images, respectively.

\begin{figure*}[tb]
\centering
  \begin{subfigure}[b]{0.28\textwidth}    \includegraphics[height=0.75\textwidth,width=\textwidth]{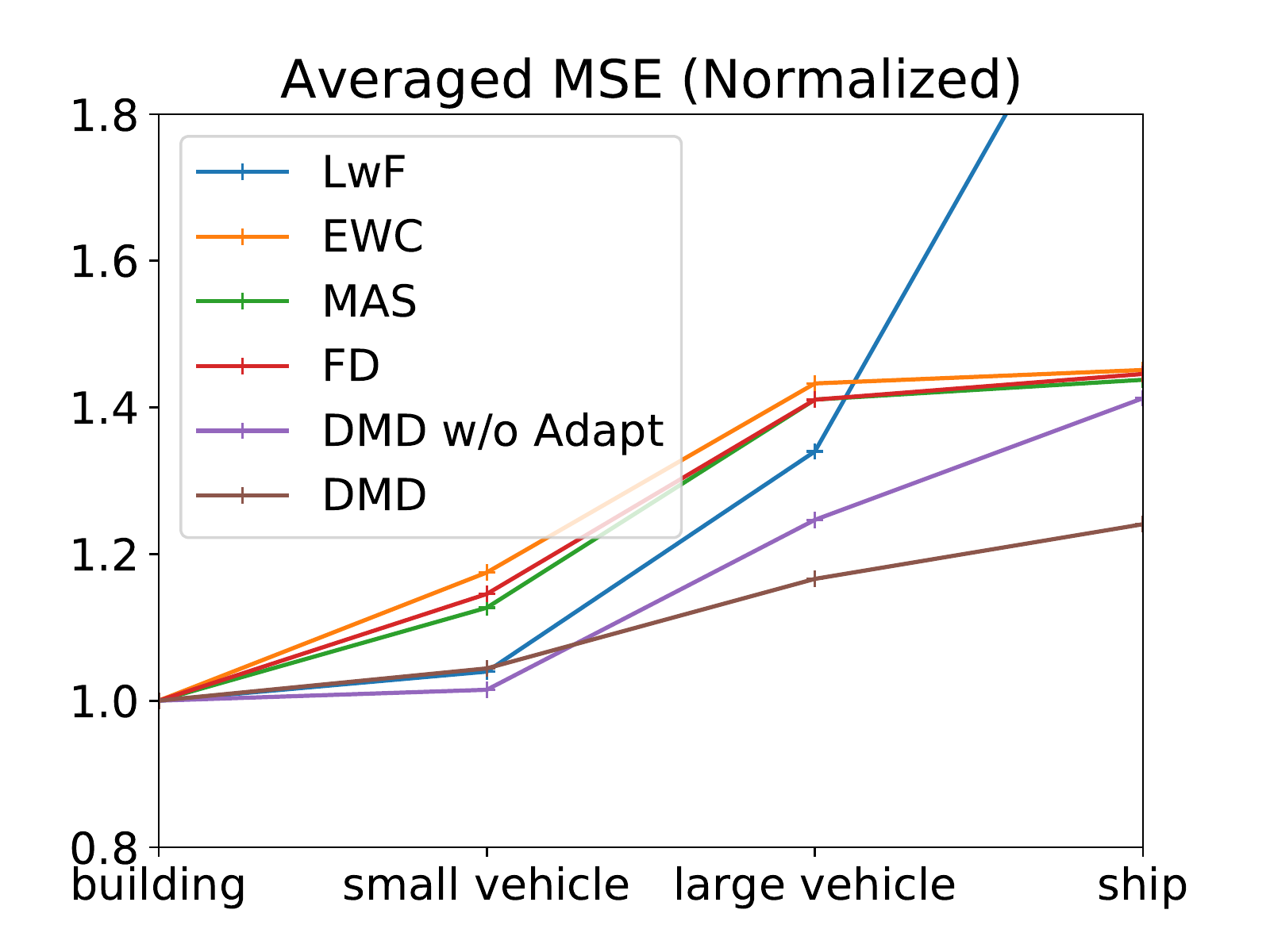}
  \end{subfigure}
  \begin{subfigure}[b]{0.28\textwidth}    \includegraphics[height=0.75\textwidth,width=\textwidth]{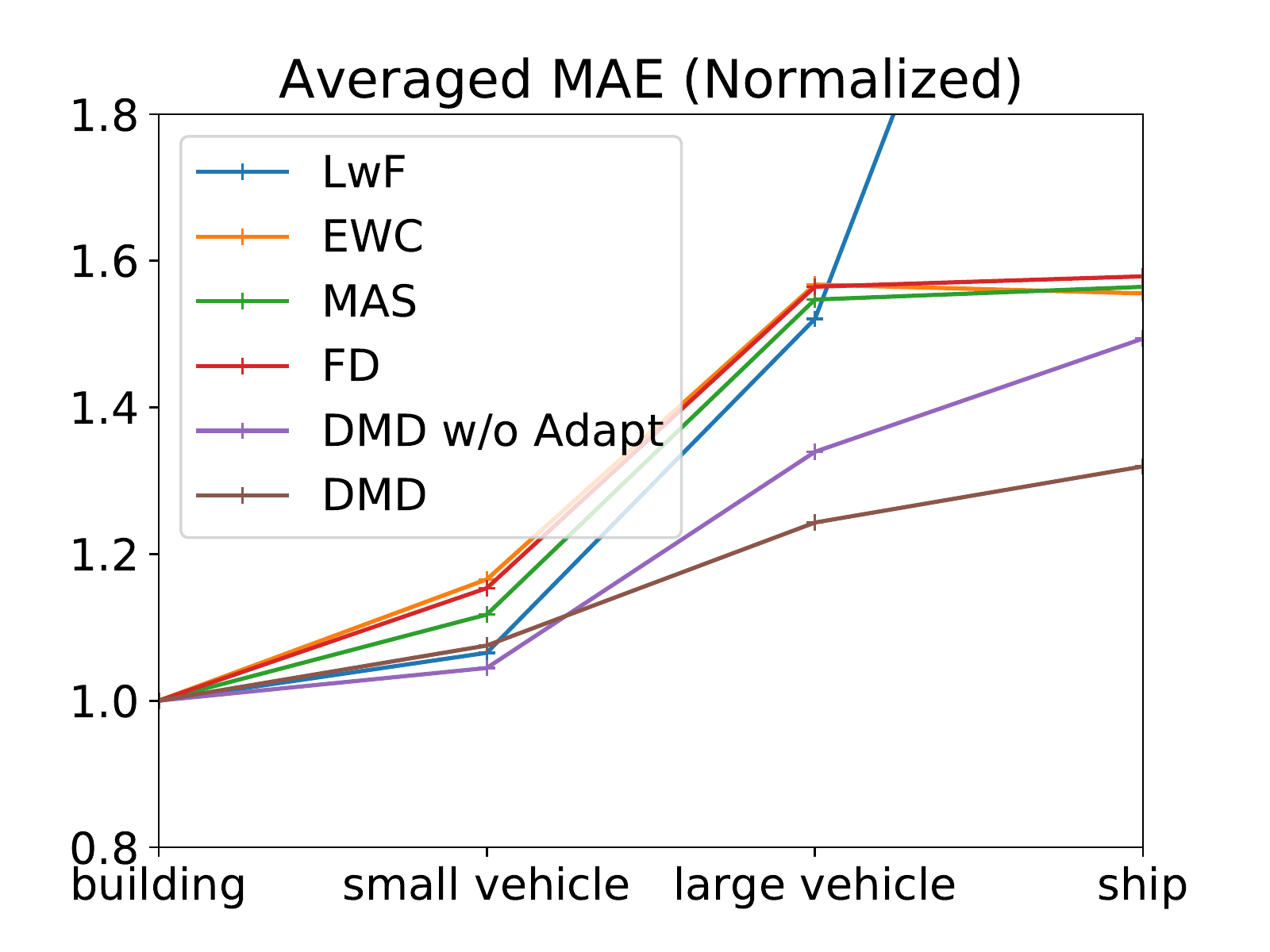}
  \end{subfigure}
  \begin{subfigure}[b]{0.28\textwidth}    \includegraphics[height=0.75\textwidth,width=\textwidth]{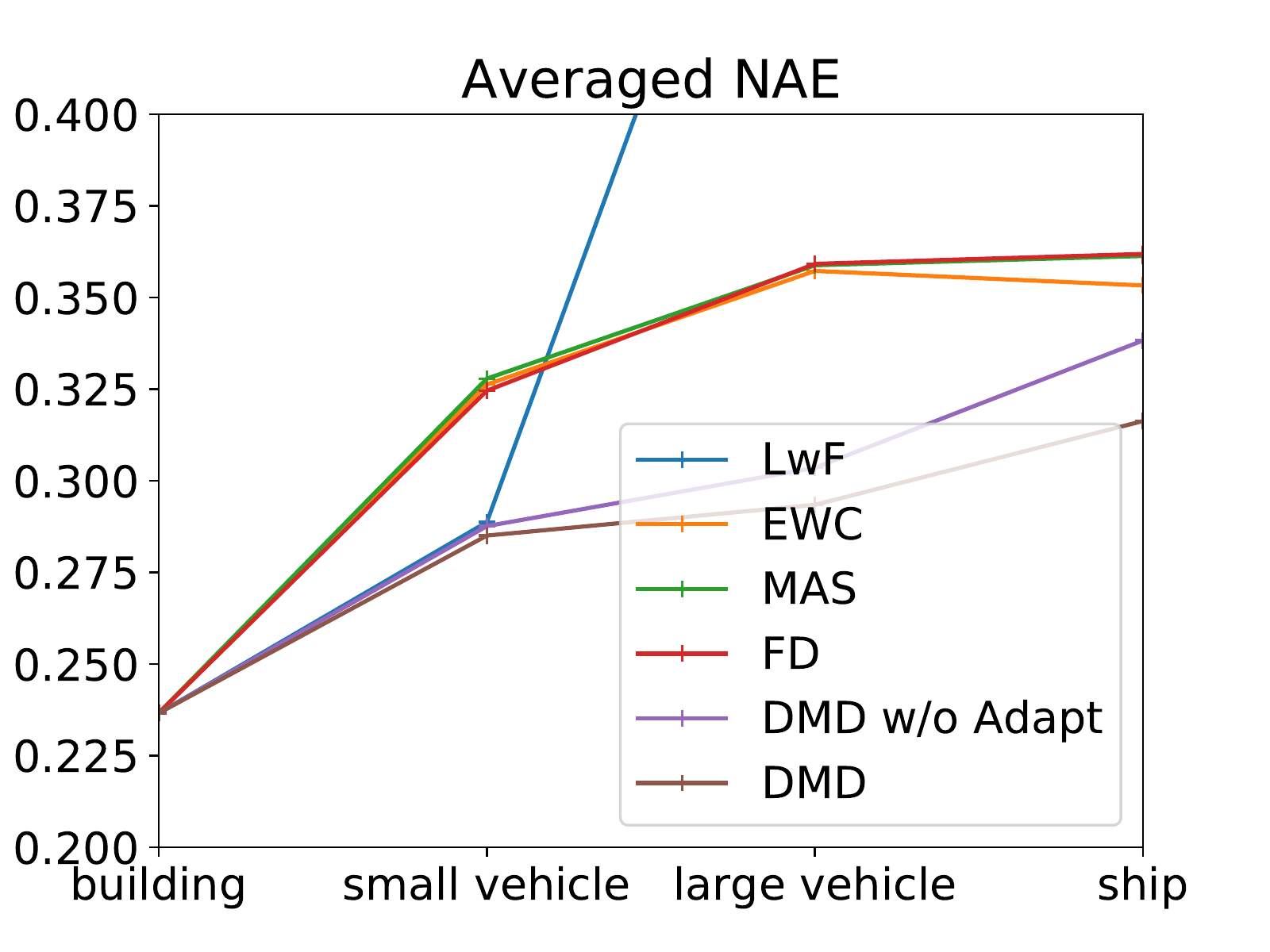}
  \end{subfigure}
  \caption{Result for RSOC (satellite) Dataset. The performances are evaluated after training of each task. We report the averaged value for all the previously seen tasks. The value is normalized to remove the dataset-scale. Lower value indicate better performance. }\label{fig:counting_result_RSOC}
\end{figure*}

\paragraph{Evaluation Metric.} Following previous methods, we use rooted Mean Squared Error (MSE), Mean Absolute Errors (MAE) and mean Normalized Absolute Errors (NAE) as metric to evaluate the performance of the model.

Mean Squared Error (MSE) is defined as:
\begin{equation}
    \text{MSE} = \frac{1}{N} \sum_{i=1}^N \left\Vert \hat y - y\right\Vert_2,
\end{equation}
where $\hat y$ is the predicted count number, $y$ is the ground truth count number and $N$ is the size for the testset.

Mean Absolute Errors (MAE) is defined as:
\begin{equation}
    \text{MAE} = \frac{1}{N} \sum_{i=1}^N \left\Vert \hat y - y\right\Vert_1, 
\end{equation}
and mean Normalized Absolute Errors (NAE) is defined as:
\begin{equation}
    \text{NAE} = \frac{1}{N} \sum_{i=1}^N \frac{\left\Vert \hat y - y\right\Vert_1}{y}.
\end{equation}
When evaluating the average performance of all the dataset, we  use NAE because its values can be compared over datasets which have varying number of objects in them.

\begin{figure*}[t]
\centering
  \begin{subfigure}[b]{0.28\textwidth}    \includegraphics[height=0.75\textwidth,width=\textwidth]{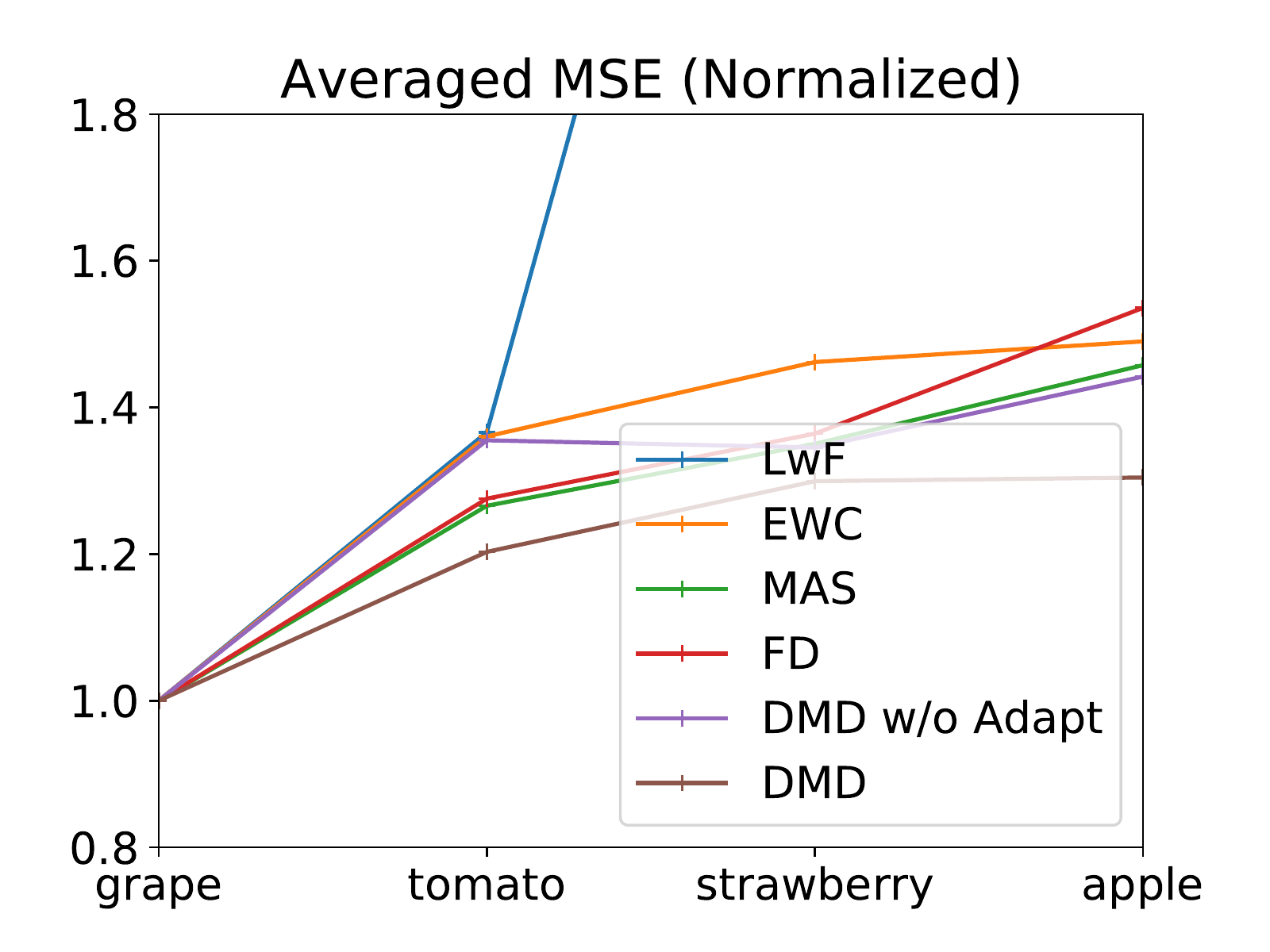}
  \end{subfigure}
  \begin{subfigure}[b]{0.28\textwidth}    \includegraphics[height=0.75\textwidth,width=\textwidth]{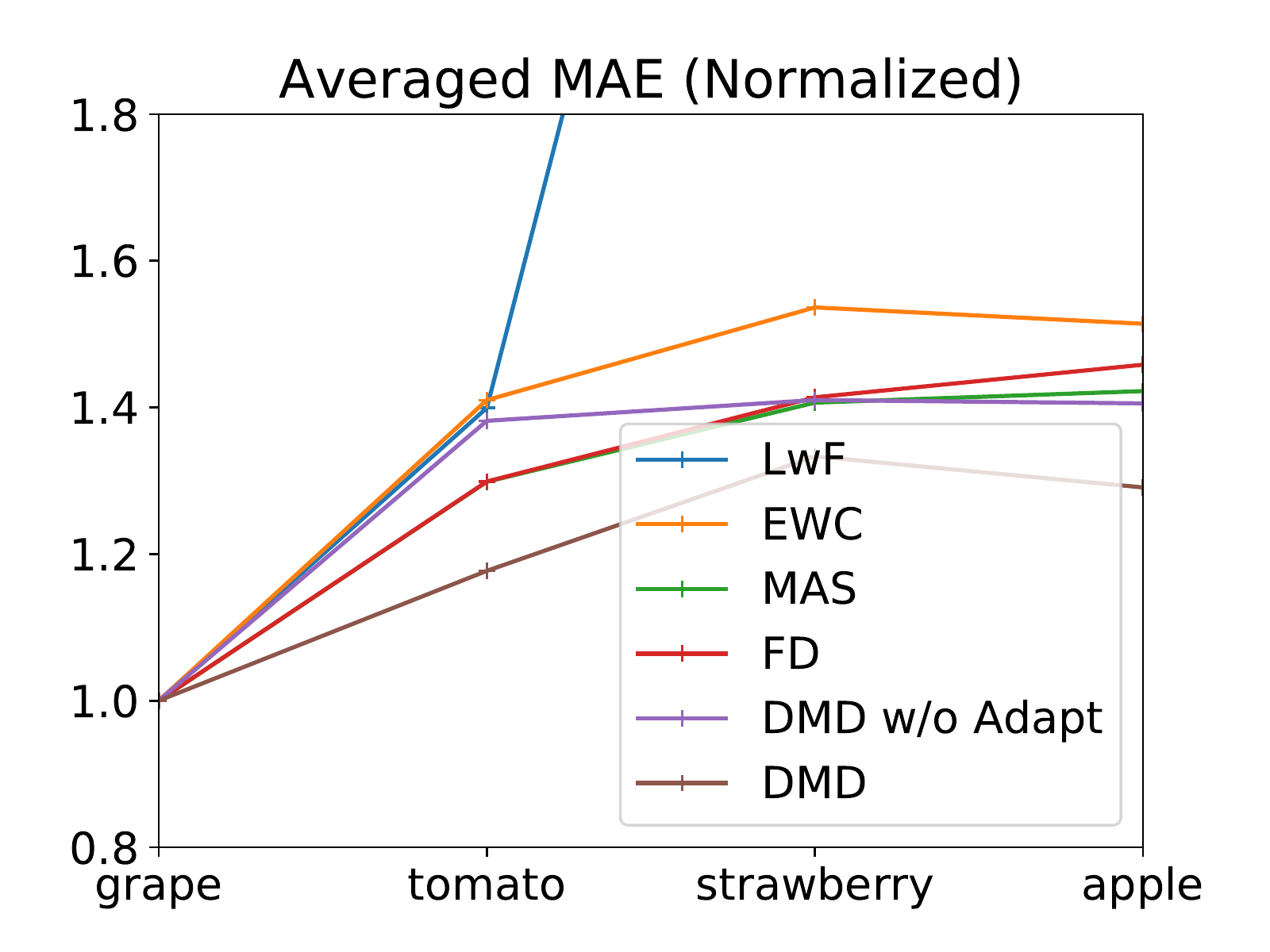}
  \end{subfigure}
  \begin{subfigure}[b]{0.28\textwidth}    \includegraphics[height=0.75\textwidth,width=\textwidth]{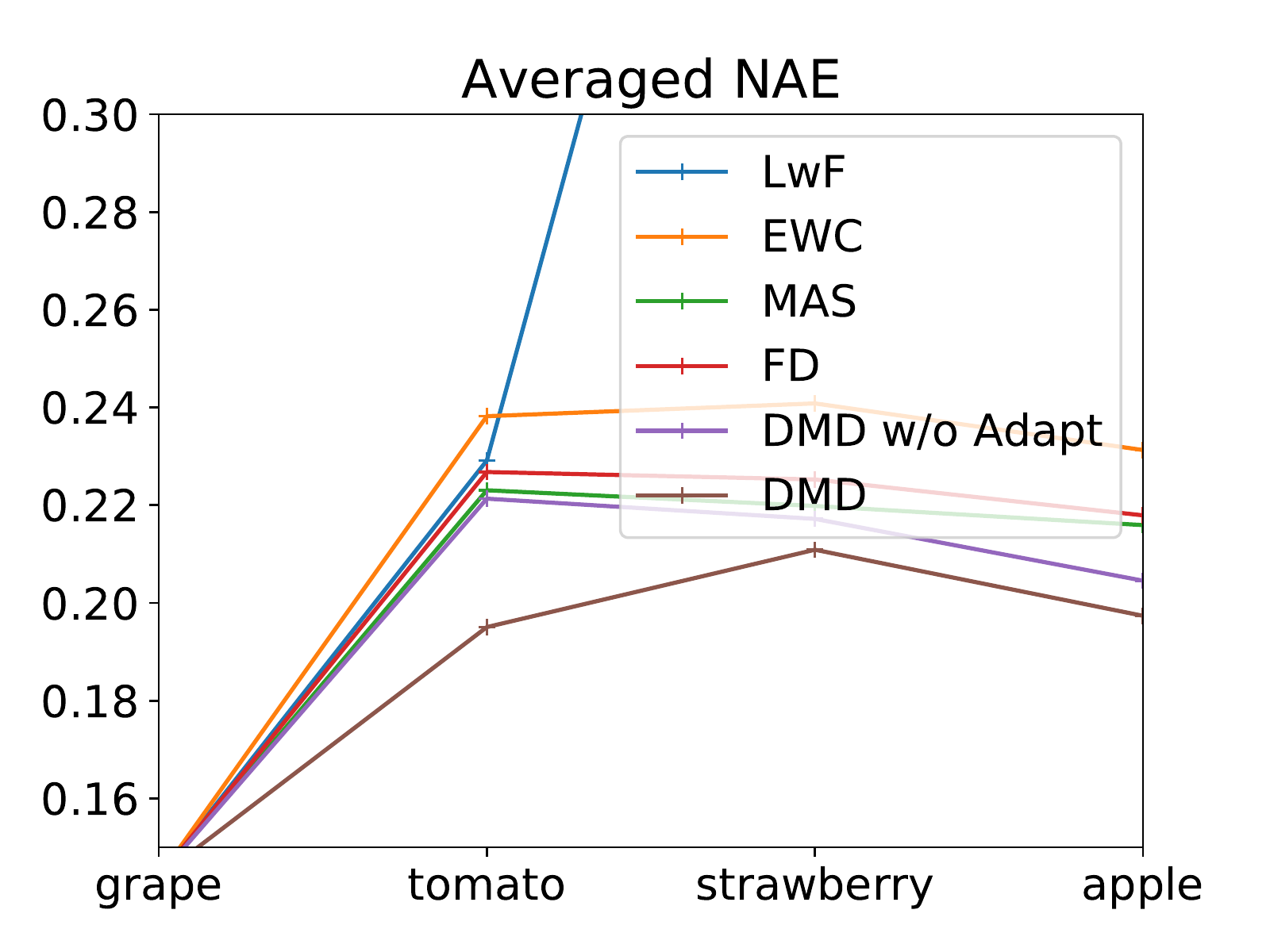}
  \end{subfigure}

  \begin{subfigure}[b]{0.28\textwidth}    \includegraphics[height=0.75\textwidth,width=\textwidth]{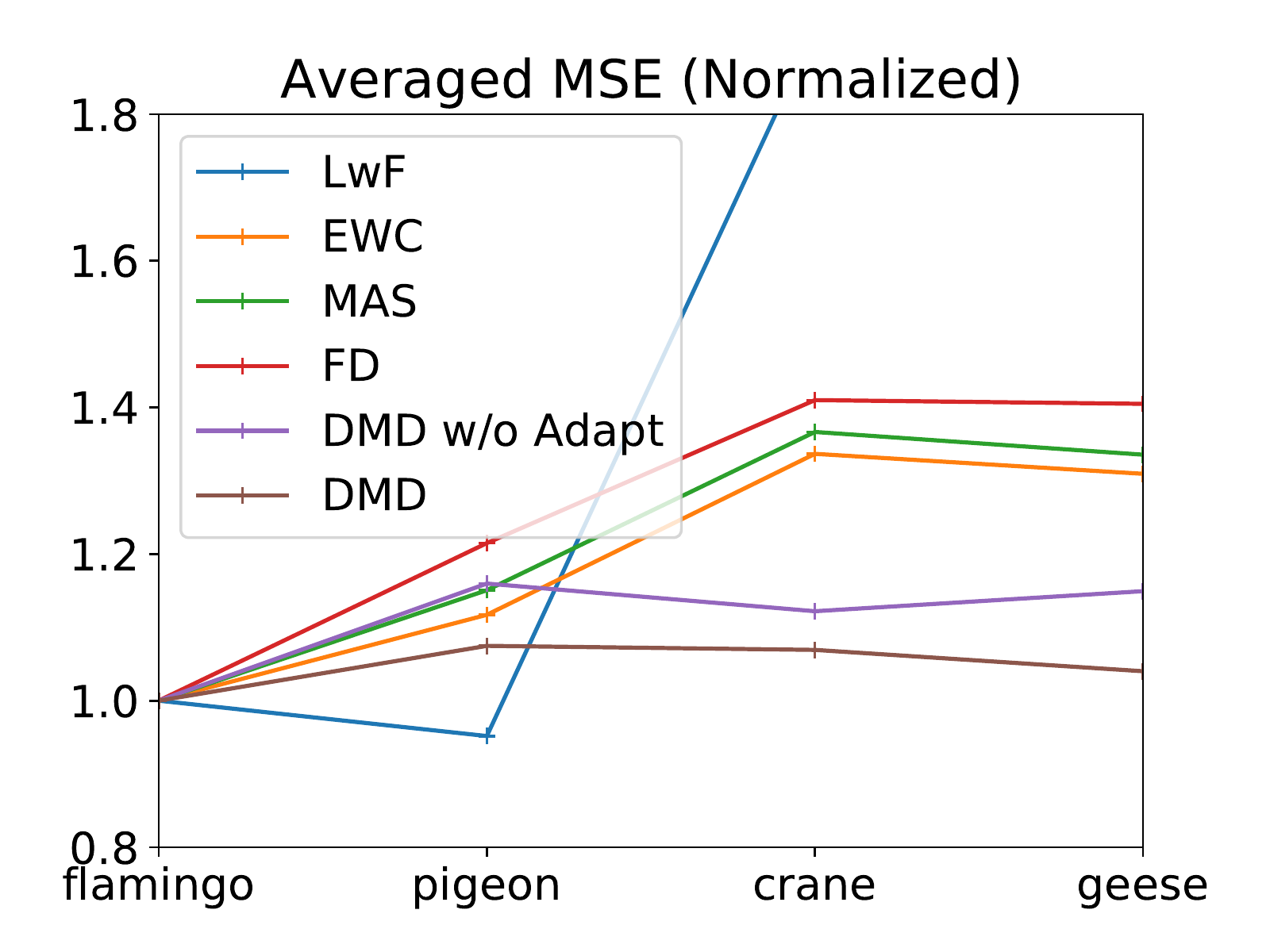}
  \end{subfigure}
  \begin{subfigure}[b]{0.28\textwidth}    \includegraphics[height=0.75\textwidth,width=\textwidth]{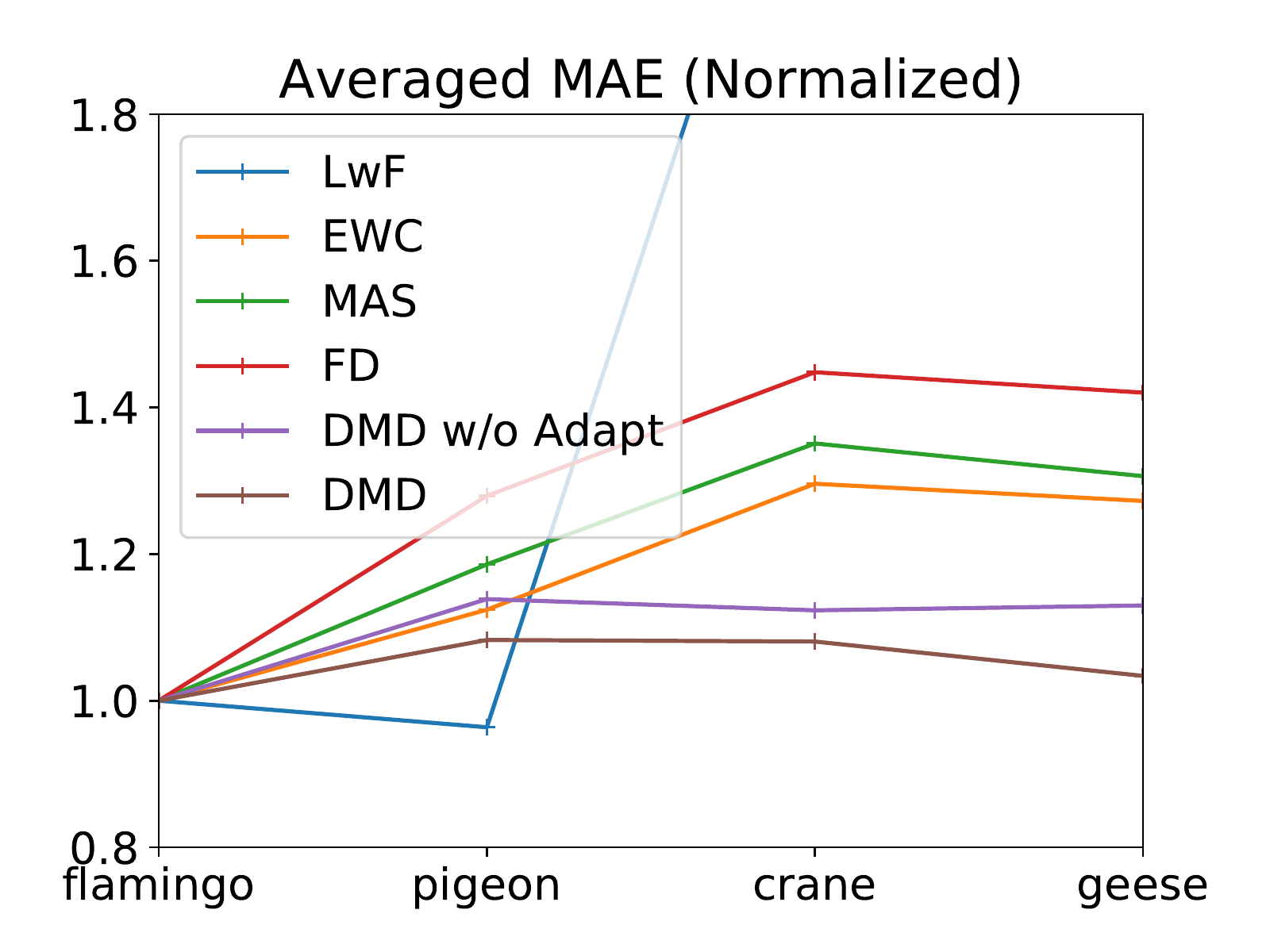}
  \end{subfigure}
  \begin{subfigure}[b]{0.28\textwidth}    \includegraphics[height=0.75\textwidth,width=\textwidth]{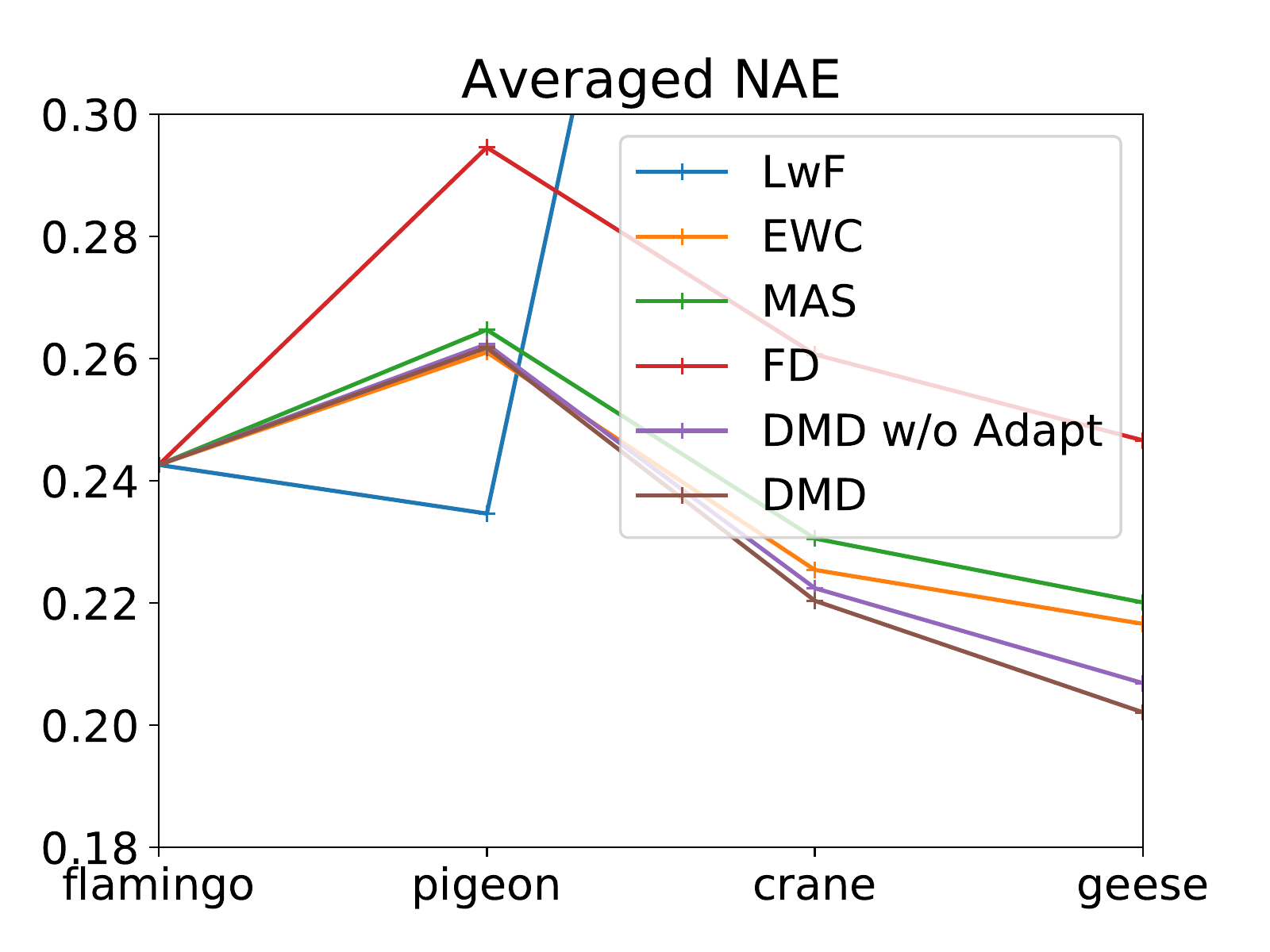}
  \end{subfigure}
  \caption{Averaged performance for FSC147 Dataset. The performances are evaluated after training of each task. We report the averaged value for all the previously seen tasks. The value is normalized to remove the dataset-scale. Lower value indicate better performance.}
  \label{fig:counting_result_FSC}
\end{figure*}

\subsection{Implementation Details}
Our implementation is based on the official code of DM-Count~\cite{dmcount2020wang}, 
and we follow its setting for training on a single dataset. 
The feature extractor is the convolutional layers of VGG19 with 512 output channels. The counter head contains of two $3 \times 3$ convolutional layers with 256 and 128 output channels respectively and a $1 \times 1$ convolutional layers with 128 output channels. 
The adapter is a one-layer $1 \times 1$ convolutional layer with the same number of channels. We train the model with the  Adam optimizer, using batch size 10, learning rate 1e-5, weight-decay 1e-4 and beta 0.9 and 0.999. For each stage, we train the model for 1000 epochs and for the next stage the training is started from the previous model. 
The hyperparameters $\lambda = 100$ for RSOC dataset and $\lambda = 10$ for both FSC-fruits and FSC-birds.

\begin{table*}[ht]
\begin{center}
\scalebox{0.6}{
\begin{tabular}{|c|ccc|ccc|ccc|ccc|c|}
\hline

\textbf{Dataset:} & \multicolumn{3}{|c|}{\textbf{grapes}}  & \multicolumn{3}{|c|}{\textbf{tomatoes}} & \multicolumn{3}{|c|}{\textbf{strawberries}}& \multicolumn{3}{|c|}{\textbf{apples}} & \textbf{Avg} \\

\textbf{Metric:} & MSE & MAE & NMAE & MSE & MAE & NMAE & MSE & MAE & NMAE & MSE & MAE & NMAE & NMAE \\
\hline
DMD w/o Adapt(1) & 24.35 & 15.85 & 0.200 & 17.15 & 13.38 & 0.302 & 18.82 & 14.70 & 0.266 & 31.26 & 16.47 & 0.195 & 0.241 \\
DMD w/o Adapt(3) & 25.35 & 18.28 & 0.240 & 14.29 & 11.50 & 0.260 & 16.09 & 11.25 & 0.177 & 23.96 & 12.33 & 0.141 & 0.207 \\
DMD(1) & 22.82 & 15.38 & 0.196 & 15.40 & 11.70 & 0.266 & 17.76 & 13.52 & 0.243 & 32.43 & 17.01 & 0.192 & 0.224 \\
DMD(3) & 26.63 & 18.83 & 0.250 & 11.35 & 8.86 & 0.221 & 16.27 & 11.09 & 0.178 & 18.56 & 10.84 & 0.140 & 0.197 \\
\hline

\end{tabular}
}
\end{center}
\caption{Ablation of varying number of layers in the counter head on the FSC-fruits sequence.}
\label{tab:counting_ablation_layer}
\end{table*}

\begin{table*}[h]
\begin{center}
\scalebox{0.6}{
\begin{tabular}{|c|ccc|ccc|ccc|ccc|c|}
\hline

\textbf{Dataset:} & \multicolumn{3}{|c|}{\textbf{grapes}}  & \multicolumn{3}{|c|}{\textbf{tomatoes}} & \multicolumn{3}{|c|}{\textbf{strawberries}}& \multicolumn{3}{|c|}{\textbf{apples}} & \textbf{Avg} \\

\textbf{Metric:} & MSE & MAE & NMAE & MSE & MAE & NMAE & MSE & MAE & NMAE & MSE & MAE & NMAE & NMAE \\
\hline
DMD w/o Adapt(10) & 25.35 & 18.28 & 0.240 & 14.29 & 11.50 & 0.260 & 16.09 & 11.25 & 0.177 & 23.96 & 12.33 & 0.141 & 0.207 \\
DMD w/o Adapt(100) & 22.47 & 15.00 & 0.190 & 14.98 & 11.25 & 0.274 & 18.14 & 12.35 & 0.210 & 23.65 & 14.23 & 0.194 & 0.217 \\
DMD(10) & 26.63 & 18.83 & 0.250 & 11.35 & 8.86 & 0.221 & 16.27 & 11.09 & 0.178 & 18.56 & 10.84 & 0.140 & 0.197 \\
DMD(100) & 22.03 & 15.09 & 0.193 & 14.93 & 11.26 & 0.272 & 17.56 & 11.69 & 0.196 & 27.44 & 15.50 & 0.190 & 0.213 \\
\hline

\end{tabular}
}
\end{center}
\caption{Ablation of different regularization hyperparameter for regularization on FSC-fruits.}
\label{tab:counting_ablation_lambda}
\end{table*}

\subsection{Results on satellite images}
For satellite images, we train our model with four classes \emph{buildings}, \emph{small vehicles}, \emph{large vehicles} and \emph{ships} in sequence from the RSOC dataset. 
Table \ref{tab:counting_table_rsoc} shows the performance at the end of the incremental learning process,  after training all four classes.  
The performance is evaluated in three metric: MSE, MAE and NMAE. Smaller values indicates better performance.
The average performance of the four classes is evaluated with NMAE because of dataset-scale invariance of the NMAE metric.

Finetuning (FT) achieves the best performance on the last task and worst on the first task, as expected. 
Feature Distillation (FD), EWC\cite{ewc2017kirkpatrick} and MAS\cite{mas2018} show a similar pattern: they are good at remembering the first task, but have difficulties to learn subsequent tasks. 
However, they also often perform good in the last task. This might be because the \emph{ships} class is more similar to the first task of \emph{buildings} when comparing to the middle \emph{vehicle} tasks.
LwF\cite{lwf2017li} performs good on the first task. But it fails in the second and third task due to its very flexible counter head. 

Our method DMD w/o Adapt improved the result compared with existing methods.
The good performance in the second and the third task shows that it can learn new task while not forgetting the previous one. 
After adding the adaptor for the feature extractor, our method DMD further improved the performance, especially on the last task. In conclusion, the proposed density map distillation obtains around a 4\% improvement over the best weight regularization method (EWC). 

Figure~\ref{fig:counting_result_RSOC} presents additional results, including the averaged MSE and MAE after learning each task, in addition to Table~\ref{tab:counting_table_rsoc}.
For example, the scores reported at task 2 in the graph  are the average of normalized MSE obtained on \emph{building} and \emph{small vehicle}) based on the network after training task 2. In the figure, we can observe that  the parameter regularization methods EWC and MAS significantly outperform the FT baseline. Next, we observe that our method DMD w/o Adapt obtains significantly better results, especially for averaged NAE. Next, we see that for only two tasks, the proposed DMD method does perform similarly to DMD w/o Adapt. However, for more tasks, DMD does significantly better, and outperforms all methods after four tasks.

\subsection{Results of counting fruits and birds}

We consider two incremental learning sequences based on the FSC147 dataset. The first one, FSC-fruits, contains the following tasks \emph{grapes}, \emph{tomatoes}, \emph{strawberries}, and \emph{apples}. The second one, FSC-birds, considers the consecutive tasks of  \emph{flamingos}, \emph{pigeons}, \emph{cranes}, and \emph{geese}. Table~\ref{tab:counting_table_FSC} summarize the results on FSC-fruits and FSC-birds.

Similar to the result in RSOC dataset, Finetuning (FT) achieves the best performance on the last task and forgets previous tasks. 
LwF~\cite{lwf2017li} gives relatively good result in the first and the last task, but failed in the second and third task. 
MAS~\cite{mas2018} and EWC~\cite{ewc2017kirkpatrick} give the best result in the first task in FSC-fruits and FSC-birds respectively, but they fail to learn new tasks. Feature Distillation (FD) also performs similarly. FD and MAS work slightly better in FSC-fruits, and EWC works better in FSC-birds.

Our method DMD w/o Adapt improves the result over the above-mentioned methods. Especially, it gets better performance in the new tasks, on both the FSC-fruit and FSC-bird sequence.
DMD further improves the result than DMD w/o Adapt, with the feature translation by the adaptor. In FSC-fruits, the performance drops slightly in the first task \emph{grapes} and improves by a large margin in the second task \emph{tomatoes}, compared with DMD w/o Adapt. In FSC-birds, the performance improves in both \emph{pigeons} (second) and \emph{geese} (last) tasks. 

Figure \ref{fig:counting_result_FSC} shows the averaged performance after training of each task. In FSC-fruits, our method DMD outperforms other methods. In FSC-birds, both DMD and DMD w/o Adapt outperform other existing method with a large margin after the third task.

\subsection{Ablation Study}
\minisection{Layers of the counter head.}
We study the effect of using different layers for the counter head in FSC-fruits benchmark. For the comparison, the size of the total network is fixed, so to increase the size of the counter head means that we move few layers from the feature extractor to the counter head. Table~\ref{tab:counting_ablation_layer} shows the result of our methods, DMD w/o Adapt and DMD, with 1 or 3 counter head layers. It shows that compared with 1 layer, using 3 layers for the counter head achieves a better performance on newer tasks and a better overall performance.

\minisection{Hyper parameter for regularization.}
We study the effect of the hyperparameter for regularization in FSC (fruits) benchmark. 
Table~\ref{tab:counting_ablation_layer} shows the result of  DMD w/o Adapt and DMD. With higher regularization, the performance for the latter tasks drop because it is too rigid for learning the new task and the model remembers the first task better.

\section{Conclusions}
We studied incremental learning for the object counting problem. The main challenge is to prevent forgetting while learning to count new object categories for new tasks. 
We propose an exemplar-free method, called Density Map Distillation (DMD). For counting each object, we train a new counter head and all tasks share a feature extractor. We propose to fix the task counter and apply a distillation loss computed with new data on the output of the old counter head. To adapt the changed feature extractor for the fixed counter head, we introduced an adaptor to project the new output feature to the old one. Experiments shows that our method DMD w/o Adapt outperforms those methods adapted from continual learning for classification problems. And with the adaptor, our DMD further improves the performance. 

\noindent \textbf{Acknowledgement.} 
We acknowledge the support of the Spanish Government funding for projects PID2019-104174GB-I00 and TED2021-132513B-I00.
{\small
\bibliographystyle{ieee_fullname}
\bibliography{mybib}
}

\end{document}